\documentclass[review]{elsarticle}

\usepackage{lineno,hyperref}
\modulolinenumbers[5]

\journal{Pattern Recognition}

\usepackage{graphicx}
\usepackage{multirow}
\usepackage{epstopdf}
\usepackage{amsmath}
\usepackage{amssymb}
\usepackage{amsfonts}
\usepackage{color}
\usepackage{subfigure}
\usepackage{verbatim}
\usepackage{adjustbox,lipsum}
\usepackage{dsfont}
\usepackage{booktabs}
\usepackage{multirow}
\usepackage{array}

\usepackage{amssymb}

\usepackage[figuresright]{rotating}

\usepackage{graphicx}
\usepackage{multirow}
\usepackage{epstopdf}
\usepackage{times}
\usepackage{amsmath}
\usepackage{amssymb}
\usepackage{amsfonts}
\usepackage{color}
\usepackage{subfigure}
\usepackage{hyperref}

\usepackage{algorithm}
\usepackage{algorithmic}
\usepackage{rotating}

\usepackage{adjustbox,lipsum}
\usepackage{dsfont}
\usepackage{booktabs}
\usepackage{multirow}
\usepackage{array}

\usepackage{pifont}

\newcommand{\xmark}{\ding{55}}%

\newcommand{\etal}{{\em et al.\,}}       
\newcommand{\eg}{{\em e.g.}}           
\newcommand{\ie}{{\em i.e.}}           

\begin{document}

\begin{frontmatter}




\title{Unsupervised Generalizable Multi-source Person Re-identification: A Domain-specific Adaptive Framework}


\author[seu]{Lei Qi}
\ead{qilei@seu.edu.cn}
\author[seu]{Jiaqi Liu}
\ead{liu\_jq@seu.edu.cn}
\author[uow]{Lei Wang}
\ead{leiw@uow.edu.au}
\author[nju]{Yinghuan Shi}
\ead{syh@nju.edu.cn}
\author[seu]{Xin Geng\corref{cor3}}
\ead{xgeng@seu.edu.cn}

\cortext[cor3]{Corresponding author: Xin Geng.}

\address[seu]{School of Computer Science and Engineering, Southeast University, Nanjing, China}
\address[uow]{School of Computing and Information Technology, University of Wollongong, Wollongong, Australia}
\address[nju]{State Key Laboratory for Novel Software Technology, Nanjing University, Nanjing, China}

\begin{abstract}
Domain generalization (DG) has attracted much attention in person re-identification (ReID) recently. It aims to make a model trained on multiple source domains generalize to an unseen target domain. Although achieving promising progress, existing methods usually need the source domains to be labeled, which could be a significant burden for practical ReID tasks. In this paper, we turn to investigate ``unsupervised'' domain generalization for ReID, by assuming that no label is available for any source domains.  To address this challenging setting, we propose a simple and efficient domain-specific adaptive framework, and realize it with an adaptive normalization module designed upon the batch and instance normalization techniques.  In doing so, we successfully yield reliable pseudo-labels to implement training and also enhance the domain generalization capability of the model as required. In addition, we show that our framework can even be applied to improve person ReID under the settings of supervised domain generalization and unsupervised domain adaptation, demonstrating competitive performance with respect to relevant methods.  Extensive experimental study on benchmark datasets is conducted to validate the proposed framework. A significance of our work lies in that it shows the potential of unsupervised domain generalization for person ReID and sets a strong baseline for the further research on this topic. The code is available at \href{https://github.com/Qi5Lei/DSAF}{https://github.com/Qi5Lei/DSAF}.
\end{abstract}


\end{frontmatter}
\section{Introduction}
\label{sec:I}
Person re-identification (ReID) is to match different images of one identity from the non-overlapping cameras in a video-surveillance system. The key is to obtain the discriminative feature to distinguish the same or different identities~\cite{arxiv20reidsurvey,qi2021adversarial,zheng2016person,DBLP:journals/tcsv/QiWHSG20}. The challenges of ReID involve the variation on views, illumination, background, resolution, etc. of different cameras. Many methods have been developed for this task. Thanks to the power of deep learning, person ReID has been well addressed in the supervised case~\cite{DBLP:conf/eccv/SunZYTW18,DBLP:conf/iccv/DaiCGZT19,DBLP:conf/iccv/ZhouYCX19,DBLP:conf/cvpr/ZhengKWR19,tao2017deep,DBLP:journals/pr/SunLCZZ21,DBLP:journals/pr/WuTLYC21}.

Recently, person ReID has been expanded to consider other settings. Unsupervised domain adaptation (UDA) methods are developed~\cite{DBLP:conf/iccv/QiWHZSG19,DBLP:conf/iccv/WuZL19,DBLP:conf/cvpr/YuZWGGL19,DBLP:conf/cvpr/ZhaiLYSCJ020,DBLP:conf/eccv/BakCL18,DBLP:journals/tip/FengCHSLC21,DBLP:journals/tifs/LiCTYQ21,DBLP:conf/cvpr/PengXWPGHT16}, with the goal of making a ReID model trained on a labelled source domain work effectively on a given target domain. Also, domain generalization (DG) has attracted much attention in person ReID~\cite{DBLP:conf/cvpr/SongYSXH19,DBLP:conf/cvpr/ChoiKJPK21,DBLP:conf/cvpr/DaiLLTD21}. It aims to make a model trained on multiple labeled source domains generalize to an unseen target domain. Both settings, particularly DG, considerably expand the application scope of person ReID in practice. However, existing UDA and DG methods still require source domains to be labeled in advance in order to train the model. It is well known that obtaining label information for person ReID is labor-expensive and time-consuming. 


\begin{figure}
\centering
\subfigure[Typical setting (Supervised)]{
\includegraphics[width=5cm]{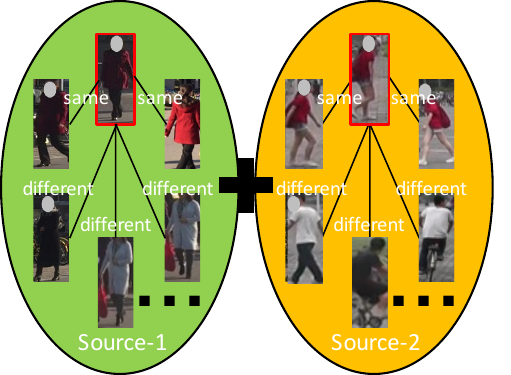}
} \label{fig:short-a}
\subfigure[Our setting (Unsupervised)]{
\includegraphics[width=5cm]{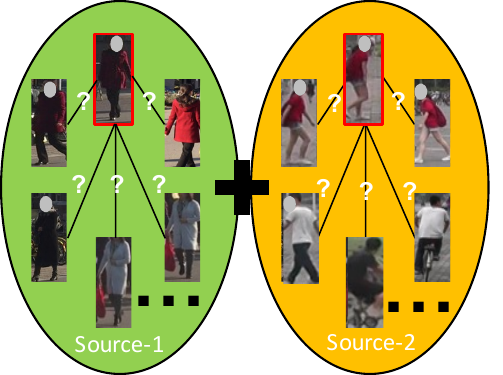}
}\label{fig:short-b}
 \caption{Comparison between the settings of typical DG-ReID and our unsupervised DG-ReID. As seen, in our task, all source domains have no label information in the training stage. Note that we use two unlabeled source domains just as an example here. The testing domain is unseen in the two tasks, as shown in Tab.~\ref{tab10}.}
\label{fig04}
\end{figure}


In this paper, we attempt a more challenging task, named unsupervised domain generalization ReID (UDG-ReID), \ie, unsupervised generalizable multi-source person re-identification. It is assumed that all label information of source domains becomes unavailable in the training stage, while the target domain remains unseen during training. Compared to the typical DG-ReID, 
our setting, as shown in Fig.~\ref{fig04}, removes the cost of data labeling in real-world applications. Besides, different from UDA, our task is now to exploit multiple unlabeled source domains and enhance the generalization ability of the model in an unseen domain, and is therefore more difficult than the UDA-ReID case. Tab.~\ref{tab10} highlights the key differences among these tasks.

For the proposed UDG-ReID task, two issues have to be addressed. Firstly, compared to UDA-ReID, which only needs to generate pseudo-labels for one target domain, we need to do this for multiple unlabeled source domains during the training course. Particularly, since there exists domain-shift between different source domains, the generation of pseudo-labels for unlabeled data therein will unavoidably be disturbed. In this case, the common practice of using a shared network path to extract the features for all source domains will not be appropriate anymore, as will be shown in Fig.~\ref{fig02} of Sec. \ref{sce:DSAF} shortly. Secondly, in the training stage, we still need to attend to the generalization ability of the model with respect to unseen target domain, as existing DG methods do. 
When there is a significant domain gap between source and target domains, the model trained on the source domains will not work well in the target domain. 

\begin{table}[t]
  \centering
  \caption{Comparison of different ReID tasks. ``ASL'' and ``ATD'' indicate the ``available source labels'' and the ``available target domain'' during training, respectively.}
    \begin{tabular}{c|c|c}
    \toprule
    Task  & ASL (Y/N) & ATD (Y/N)\\
    \midrule
    UDA-ReID   & $\checkmark$     & $\checkmark$ \\
    UDAw/oSL-ReID (ours) & \xmark     & $\checkmark$ \\
    \midrule
    DG-ReID    & $\checkmark$     & \xmark  \\
    UDG-ReID (ours) & \xmark     & \xmark  \\
    \bottomrule
    \end{tabular}%
  \label{tab10}%
\end{table}%

In this paper, we address the above two issues in a unified framework. To this end, a domain-specific adaptive framework (DSAF) is proposed for the UDG-ReID task. Specifically, to reduce the impact of domain shift on pseudo-label generation, 
we create a domain-specific network path for each of the source domains, so as to achieve more accurate pseudo-label prediction. At the same time, to obtain good generalization ability on unseen target domain, we create another path to adaptively mitigate the domain gap between the multiple source domains and the unseen target domain. Following the design, this work realizes the proposed framework via the normalization techniques in deep learning. This produces the domain-specific adaptive normalization (DSAN) which will become clear shortly. 

We conduct experiments on multiple benchmark datasets to show the potential of our method. 
Besides, we demonstrate that our method can also be adapted to supervised DG-ReID setting and outperform existing methods.
Furthermore, motivated by the proposed UDG-ReID, we introduce another new setting called UDA without source labels (UDAw/oSL), as shown in Tab.~\ref{tab10}. Although requiring no labels in source domain, with our method, it can yield competitive performance versus existing UDA-ReID methods. Our contributions are summarized as below:
      \begin{itemize}
    \item We propose an unsupervised domain generalization person ReID (UDG-ReID) task in the person ReID community. It does not need label information for source domains, and is therefore more challenging but practical than typical DG-ReID task. 
    \item We develop a simple yet effective domain-specific adaptive framework to reduce the adverse impact of domain gap across source domains during generating pseudo-labels and boost the generalization ability of the model for the unseen target domain.
    \item We evaluate our method on multiple benchmark datasets. The results show that it achieves higher accuracy than baselines on all datasets. Besides, our method adapted to supervised DG-ReID task 
    and the new task of UDA without source labels also demonstrates promising performance. 
  \end{itemize}

The rest of this paper is organized as follows.
We review some related work in Section \ref{s-related}.
The proposed method is introduced in Section \ref{s-framework}.
Experimental results and analysis are presented in Section \ref{s-experiment},
and Section \ref{s-conclusion} is conclusion.
\section{Related Work}\label{s-related}
In this section, we review some related works to our work in the following part, including supervised domain generalization (DG) person ReID and unsupervised domain adaptation (UDA) person ReID.

\subsection{Supervised DG Person ReID}
Some existing DG-ReID methods aim to learn the domain-invariant feature for any domain. 
Jia  \etal~\cite{DBLP:conf/bmvc/JiaRH19} observe that appropriate instance and feature normalization alleviates style and content variance across datasets in deep ReID models.
~Jin \etal~\cite{DBLP:conf/cvpr/JinLZ0Z20} propose to distill the identity-relevant feature from the information removed by instance normalization (IN)~\cite{ulyanov2016instance}  and restore it to the network to ensure high discrimination. 
Zhao \etal~\cite{DBLP:conf/cvpr/ZhaoZYLLLS21} develop the memory-based multi-source meta-learning framework to train a generalizable model for unseen domains. 
In~\cite{DBLP:conf/cvpr/SongYSXH19}, a novel deep ReID model termed domain-invariant mapping network is proposed, which can learn a mapping between a person image and its identity classifier. 

In addition, some methods employ the adaptive technique to make the trained model better generalize in the unseen target domain.
The method~\cite{DBLP:conf/eccv/LiaoS20} treats image matching as finding local correspondences in feature maps and constructs query-adaptive convolution kernels on the fly to
achieve local matching. 
To generalize normalization layers, this work~\cite{DBLP:conf/cvpr/ChoiKJPK21}  combines learnable batch-instance normalization layers with meta-learning. 
Differently, in~\cite{DBLP:conf/cvpr/DaiLLTD21},  a voting-based mixture mechanism is used to dynamically leverage diverse characteristics of source domains, which can better improve the generalization ability of models. 
Compared to the supervised DG person ReID, there is no available label information in the training stage in our task, \ie, we do not know the label of the source domain, as shown in Tab.~\ref{tab10}.

In our UDG-ReID task, we not only need to learn the robust model in the unseen domain, but also reduce the interference of domain gap between unlabeled source domains to produce the better pseudo-labels.

In addition, several DG methods have also obtained great success in the classification task and the semantic segmentation task~\cite{DBLP:conf/eccv/ZhouYHX20,DBLP:conf/iclr/ZhouY0X21,DBLP:conf/eccv/HuangWXH20,DBLP:conf/aaai/ZhouYHX20,Li_2021_ICCV,DBLP:journals/pr/ZhangQSG22}.
For example, Nam \etal~\cite{DBLP:conf/cvpr/NamLPYY21} propose to reduce the intrinsic style bias of CNNs to close the gap between domains. 
Considering that the Fourier phase information contains high-level semantics and is not easily affected by domain-shift, Xu \etal~\cite{DBLP:conf/cvpr/XuZ0W021} introduce a novel Fourier-based perspective for domain generalization.  
Seo \etal~\cite{DBLP:conf/eccv/SeoSKKHH20} develop a simple but effective multi-source domain generalization technique based on deep neural networks by incorporating optimized normalization layers that are specific to individual domains.
~However, most of these methods do not also need to tackle the case without the labels of all source domains.

\subsection{UDA Person ReID}
As mentioned in Sec. \ref{sec:I}, our task is also related to the UDA-ReID case. Most of the successful UDA-ReID methods use clustering-based pseudo-label prediction with representation learning and perform the two
steps in an alternating fashion~\cite{DBLP:conf/eccv/LuoSZ20,DBLP:conf/eccv/0014LZ020,DBLP:conf/eccv/ChenLL020,DBLP:conf/cvpr/LinXWY020,DBLP:journals/tmm/YangYLJXYGHG21,DBLP:journals/tip/BaiWLLD21,zhang2020improving,DBLP:journals/pr/ZhengXCHCZ22}. However, because of the domain gap, the pseudo-labels from the model trained in the source domain are not always reliable and there are incorrect labels. Most of existing methods mainly aim to alleviate the negative effects of noisy labels.

In~\cite{DBLP:conf/cvpr/YangZLCLLS21}, a dynamic and symmetric cross-entropy loss is developed to deal with noisy samples and accommodate to the change of clusters. 
Zheng \etal~\cite{DBLP:conf/aaai/ZhengLZZZ21} estimate and exploit the credibility of the assigned pseudo-label of each sample by suppressing the contribution of noisy samples.  
Differently, Ge \etal~\cite{DBLP:conf/iclr/GeCL20} propose to softly refine the pseudo-labels in the target domain by proposing an unsupervised framework to learn better features from the target domain via off-line refined hard pseudo-labels and on-line refined soft pseudo-labels in an alternative training manner. The method in~\cite{DBLP:conf/cvpr/ZhengLH0LZ21} enables the online interaction and mutual promotion of pseudo-label prediction and representation learning to better correct the noisy labels.

Besides, Xuan \etal~\cite{DBLP:conf/cvpr/XuanZ21} split the sample similarity into the intra-camera and inter-camera computations, respectively, which effectively alleviates the distribution gap among cameras to generate more reliable pseudo-labels. In particular, most of the existing pseudo-label-based methods mainly aim to produce the pseudo-label for an unsupervised domain. Differently, in our task, all domains in the training stage are unlabeled, hence we aim to alleviate the impact of the inter-domain interference during the pseudo-label generation.
In~\cite{DBLP:conf/nips/Ge0C0L20}, a novel self-paced contrastive learning framework is proposed to exploit all valuable information. 
Moreover, to use the valuable labeled data, a rectification domain-specific batch normalization module is introduced for the multi-source UDA-ReID in~\cite{DBLP:conf/cvpr/BaiWW0D21}. 

These aforementioned methods try to handle the case without the labels of the target domain. However, unlike the UDA-ReID task, in our UDG-ReID task, the target domain cannot be seen in the training stage, and all source domains do not include the label information, as shown in Tab.~\ref{tab10}.

\section{Methodology}\label{s-framework}
In this section, we firstly introduce the unsupervised domain generalization person ReID (UDG-ReID). Then, we present the proposed domain-specific adaptive framework (DSAF). Moreover, we describe the domain-specific adaptive normalization (DSAN). Finally, we give the details of the network.

\begin{figure}
\centering
\includegraphics[width=8cm]{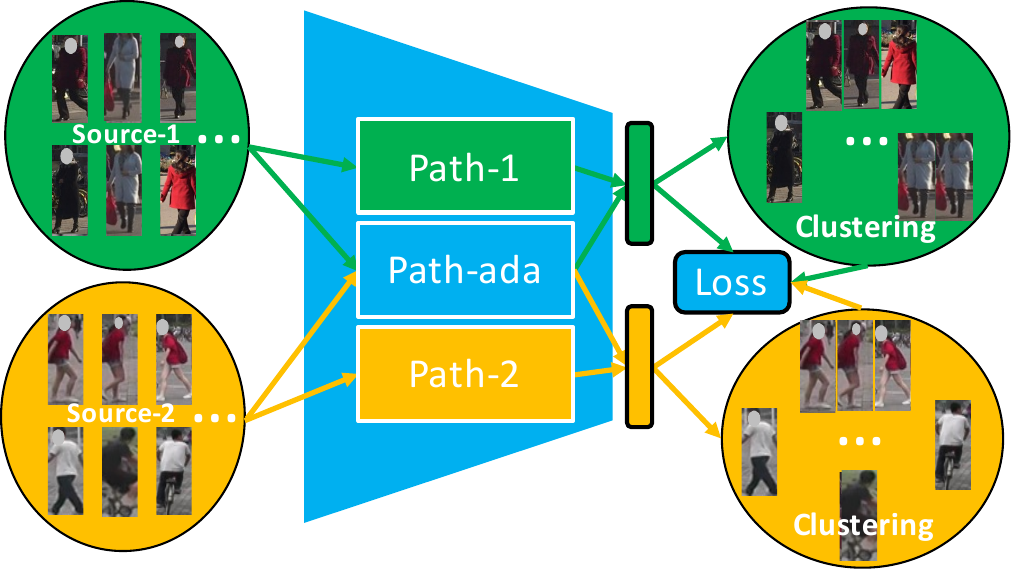}
 \caption{Overview of the proposed DSAF. Note that we take two unlabeled source domains as an example in this figure. In the framework, the independent path for each domain is to alleviate the impact of the inter-domain interference during the pseudo-label generation. The ``path-ada'' is to boost the generalization ability in the unseen target domain.}
\label{fig07}
\end{figure}

\subsection{Unsupervised Domain Generalization Person ReID}
Recently, an unsupervised domain generalization (UDG) task has been proposed for the classification task~\cite{zhang2021domain}, which can be considered as a self-training task~\cite{DBLP:conf/icml/ChenK0H20,DBLP:conf/cvpr/He0WXG20} with the domain-shift in the training samples, \ie, training a pre-trained model. It is worth noting that, the UDG task for classification needs a few labeled images in the target domain. The main reason is that the pre-trained model cannot predict which class an image belongs to, thus a few labeled images from the target domain are required to train the classifier or fine-tune the pre-trained model. Therefore, the UDG for classification is not a strict domain generalization (DG) task.  

Differently, person ReID is a metric task, which aims to learn the discriminative features to evaluate the similarity between different images, \ie, it does not require predicting which ID an image belongs to. Particularly, the current existing tasks for mitigating the requirement of labeling or collecting data in the person ReID mainly include the UDA and DG. For the UDA task, it still needs to collect data from the target domain, while the DG task requires the labeled data in the training stage. To further alleviate this issue, we propose an unsupervised domain generalization Re-ID task, where there is no label information for the training data and the labeled images of the target domain are not required during testing. In our task, assuming that we have $N_d$ domains ($\{D_1, D_2, \ldots, D_{N_d}\}$) without labels to train the network, we aim to output a generalizable model $\hat{\theta}$ to the unseen domain.

It is worth noting that the difference between our setting and domain-incremental learning~\cite{van2019three,yang2019adaptive} is the definition of the domain. In domain-incremental learning, all domains belong to the same dataset, while all classes in each domain are not overlapping, \ie, the distribution of the \textbf{output} is different. In contrast, in our setting, each domain belongs to different datasets, \ie, the distribution of the \textbf{input} is different.

\subsection{Domain-Specific Adaptive Framework}
\label{sce:DSAF}
For the proposed unsupervised domain generalization person ReID (UDG-ReID), there are two key problems as below: 1) there is no label information in all source domains, and 2) the target domain could be different from all source domains from the data-distribution perspective. Therefore, most of existing DG-ReID or UDA-ReID methods cannot be directly applied to deal with the UDG-ReID.

\begin{figure}
\centering
\subfigure[Performance]{
\includegraphics[width=5cm]{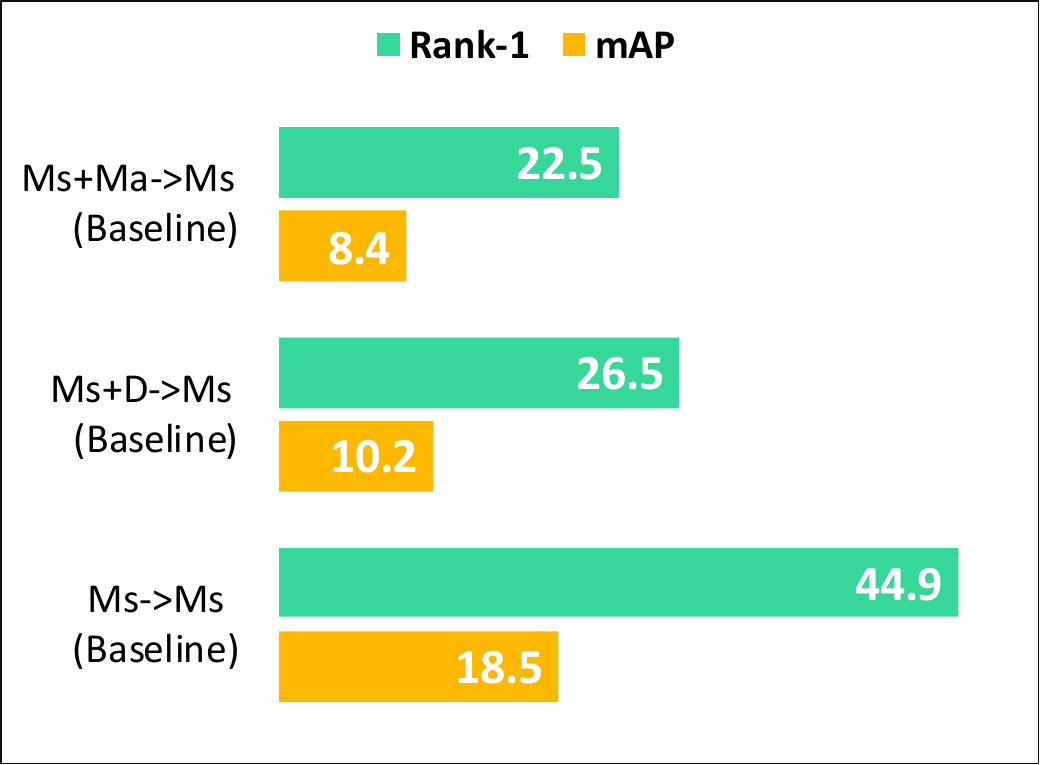}
} 
\subfigure[Clustering evaluation]{
\includegraphics[width=5cm]{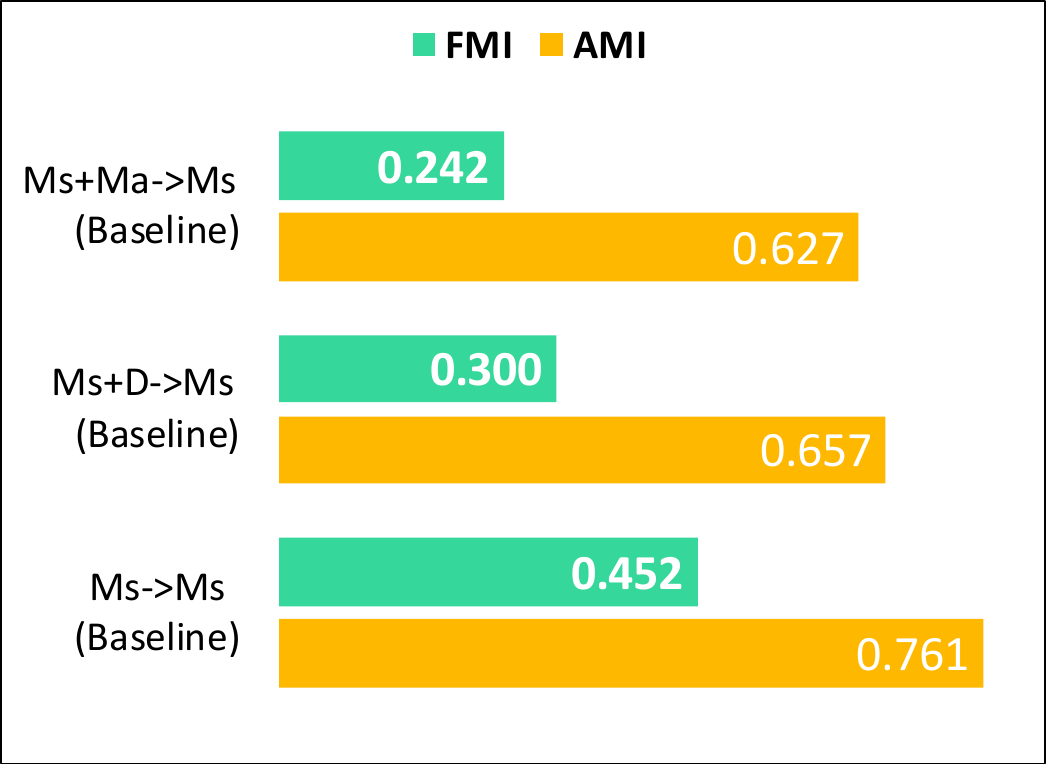}
}
 \caption{Experimental results of different settings using the raw ResNet-50, \ie, the baseline network in this paper, on MSMT17 (Ms). Note that ``Ms+Ma$\rightarrow$Ms'' is that training the model on ``Ms'' and ``Ma'' (Market-1501), and evaluating the model on ``Ms''. Besides,  ``D'' is DukeMTMC-reID in this figure. AMI (Adjusted Mutual Information) and FMI (Fowlkes-Mallows Index)~\cite{DBLP:journals/jmlr/NguyenEB10} are two clustering evaluation protocols, and the larger value is better. Note that there are no available labels in the experiment.}
\label{fig02}
\end{figure}

For the former issue, 
in the UDG-ReID task, we need to generate the pseudo-labels for all source domains, thus the discriminative feature representation is required for each unlabeled sample. However, since there is the domain gap between different source domains, if we utilize a completely shared path for all source domains in the backbone network during training, each domain will be disturbed each other during pseudo-labeling. As shown in Fig.~\ref{fig02} (a), the model trained on the MSMT17 dataset can obtain a better performance than the model jointly trained on MSMT17 and other datasets when evaluated on MSMT17. It is worth noting that we do not provide the label information for all training samples in the experiment. Besides, we report the clustering evaluation of unlabeled training data on MSMT17, as shown in Fig.~\ref{fig02} (b), and there is the similar pattern to Fig.~\ref{fig02} (a). This reveals that jointly training model using multiple unlabeled datasets (\ie, domains) reduces the robustness of features due to the domain gap between different training domains, which decreases the reliability of pseudo-labels. Therefore, for UDG-ReID, we need to deal with this issue.

For the latter issue, the generalization ability of the model is key for the unseen target domain due to the domain gap between source and target domains. Most of the existing methods aim to improve the robustness of the model by learning the domain-invariant feature or conducting the augmentation operation on the features or images. Particularly, different from the conventional DG task, we aim to simultaneously consider the discrimination ability and the generalization ability of a model in the UDG-ReID task because of no available labels in all source domains.

In this paper, we propose a domain-specific adaptive framework (DSAF) to solve the above issues, as shown in Fig.~\ref{fig07}. In this framework, we use the private path for each unlabeled source domain, which alleviates the impact of the inter-domain interference during the pseudo-label generation. Besides, we employ an adaptive path to boost the generalization ability in the unseen target domain, \ie, a sample-adaptive path in the testing stage, which can remove the impact of the domain gap between sources domains and the unseen target domain when we extract features from the unseen domain. Finally, we fuse the features from the domain-specific path and the global adaptive path to conduct the clustering operation, and then generate pseudo-labels to train the model. After each epoch, we utilize the clustering algorithm to regenerate the new pseudo-labels.

\subsection{Domain-Specific Adaptive Normalization}

\begin{figure}[t]
\centering
\includegraphics[width=7.5cm]{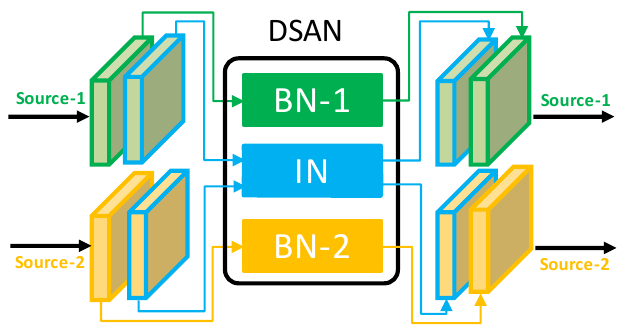}
 \caption{Illustration of the proposed DSAN. In this figure, we take two source domains as an example.}
\label{fig05}
\end{figure}

Since there is the domain gap between different domains, the statistics of batch normalization (BN) from different domains in the neural network are different. Therefore, domain-specific batch normalization (DSBN)~\cite{DBLP:conf/cvpr/ChangYSKH19} is proposed to alleviate the domain-shift between source and target domains in the classical UDA task. Besides, the instance normalization (IN) can enhance the generalization ability of the model by removing the style information with the sample-adaptive statistics in the DG task~\cite{DBLP:conf/bmvc/JiaRH19}.
In this paper, inspired by these above methods, we develop a domain-specific adaptive normalization (DSAN) to achieve the proposed DSAF from the normalization perspective, which can simultaneously mitigate the interference between different source domains during generating pseudo-labels and boost the generalization ability in the unseen target domain in the unified framework, as illustrated in Fig.~\ref{fig05}. In DSAN, DSBN and IN achieve the domain-specific path and the domain-adaptive path of DSAF in Fig.~\ref{fig07}, respectively. Inspired by IBN-Net~\cite{DBLP:conf/eccv/PanLST18}, we use a half feature maps into IN and another feature maps to conduct the domain-specific BN, and then utilize the concat operation to fuse them.
 Given an example from domain $d$, its feature maps can be defined as $f_{d} \in \mathbb{R}^{N\times C\times  H\times W}$, where $H$ and $W$ indicate spatial dimensions, and $C$ and $N$ are the number of channels and batch size. We split $f_{d}$ into two parts according to channel dimension, $f_{d} = f_{d}^1 \oplus_c f_{d}^2$, where $f_{d}^1, f_{d}^2 \in \mathbb{R}^{N\times C/2\times H\times W}$, and $ \oplus_c$ represents the concat operation based on the channel dimension. Thus, the global IN and the BN for the $d$-th domain can be formulated as:
  \begin{equation}
  {\rm IN}(f_d^1; \gamma^{in}, \beta^{in})= \gamma^{in} \frac{f_d^1-\mu_n}{\sigma_n}+\beta^{in},
  \label{eq01}
  \end{equation}
  \begin{equation}
  \begin{aligned}
  &{\rm BN_d}(f_d^2; \gamma_b^{bn}, \beta_b^{bn})= \gamma_d^{bn} \frac{f_d^2-\mu_d}{\sigma_d}+\beta_d^{bn},~~d \in \{1, ..., D\},
  \end{aligned}
    \label{eq02}
  \end{equation}
  where $\gamma^{in},\beta^{in}, \gamma_d^{bn},\beta_d^{bn} \in \mathbb{R}^{C/2}$ are learnable affine transformation parameters, and $D$ is the number of domains. $\mu_n, \sigma_n$ and $\mu_d, \sigma_d \in \mathbb{R}^{C/2}$ represent the channel-wise mean and standard deviation of IN and BN of each feature map as follows:
  
  \begin{equation}
  \mu_n=\frac{1}{HW}\sum_{h=1}^{H}\sum_{w=1}^{W}f_d^1[n,:,h,w],
  \label{eq:7}
  \end{equation}
  \begin{equation}
  \mu_d=\frac{1}{NHW}\sum_{n=1}^{N}\sum_{h=1}^{H}\sum_{w=1}^{W}f_d^2[n,:,h,w],
  \label{eq:7}
  \end{equation}
  \begin{equation}
  \sigma_n=\sqrt{\frac{1}{HW}\sum_{h=1}^{H}\sum_{w=1}^{W}(f_d^1[n,:,h,w]-\mu_n)^2 + \epsilon},
  \label{eq:8}
  \end{equation}
  \begin{equation}
  \sigma_d=\sqrt{\frac{1}{NHW}\sum_{n=1}^{N}\sum_{h=1}^{H}\sum_{w=1}^{W}(f_d^2[n,:,h,w]-\mu_d)^2 + \epsilon},
  \label{eq:8}
  \end{equation}
  where $\epsilon$ is a constant for numerical stability. 
~Therefore, our domain-specific adaptive normalization is defined as:
  \begin{equation}
    \begin{aligned}
\hat{f_d} = &DSAN(f_d) = {\rm IN}(f_d^1; \gamma^{in}, \beta^{in}) \oplus_c \\
 &{\rm BN_d}(f_d^2; \gamma_b^{bn}, \beta_b^{bn}),~~d \in \{1, ..., D\}.
   \end{aligned}
  \end{equation}

According to the above description, the forward process of the proposed DSAN is summarized in Algorithm~\ref{al01}.

\begin{algorithm}[ht]
\caption{The forward process of DSAN}
\begin{algorithmic}[1]
\STATE {\bf Input:} 
Give the $d$-th domain feature $f_d\in \mathbb{R}^{N\times C\times  H\times W}$.\\
\STATE {\bf Output:} The normalized feature $\hat{f_d}$. \\
\STATE Evenly split $f_d$ into $f_d^1$ and $f_d^2$ on channel dimension.
\STATE Conduct instance normalization for $f_d^1$ with the shared affine transformation as Eq.~(\ref{eq01}), output $\hat{f_d^1}$.
\STATE Conduct batch normalization for $f_d^2$ with the private affine transformation for the $d$-th domain as Eq.~(\ref{eq02}), output $\hat{f_d^2}$.
\STATE Concat $\hat{f_d^1}$ and $\hat{f_d^2}$ on channel dimension to generate the normalized feature $\hat{f_d}$.
\end{algorithmic}
\label{al01}
\end{algorithm}

\textit{Remark.} 
The advantage of using the DSAN to achieve the DSAF has two folds. Firstly, DSAN can simultaneously tackle the inter-domain interference during pseudo-labeling and domain gap between source domains and unseen target domain. Secondly, using the DSAN merely adds a small number of parameters in the model due to sharing all layers except for the normalization layer. 
It is worth noting that DSAN is not the unique method to implement DSAF, thus DSAF could be implemented via other methods.

Although the DSAF needs to add a new path for each new dataset, the number of the added parameter in a new path is extremely small because the method only changes the normalization layer of the baseline. Here, we show the memory size of our method (30190MB) and the baseline (30014MB) when using two datasets in the training stage. As seen, our method does not add a large memory size.
\subsection{Details of Network Framework}
\label{sec:DNF}

In this paper, we use ResNet-50~\cite{DBLP:conf/cvpr/HeZRS16} as backbone, which mainly consists of four blocks. We employ our proposed DSAN in all blocks by replacing the original BN. Besides, for generating the pseudo-labels, we utilize DBSCAN~\cite{ester1996density} to conduct the clustering after each epoch. Based on these pseudo-labels, we directly employ the cross-entropy loss ($\mathcal{L}_{cls}$) and triplet loss ($\mathcal{L}_{tri}$) with hard sample mining~\cite{DBLP:journals/tmm/LuoJGLLLG20,hermans2017defense,DBLP:conf/iccv/FuWWZSUH19} to train the model as below:

  \begin{equation}
  \mathcal{L}_{total} =  \mathcal{L}_{cls} +  \mathcal{L}_{tri}.
  \end{equation}

In the testing stage, since our method for each source domain has an independent path, the features from all paths are averaged as the final feature. 

\textit{Remark.}  Our method can also be used into the supervised DG-ReID case when given the label information of all sources. Using the domain-specific framework could extract the features of the unseen target domain from the different views because different domain-specific paths map the same sample into different spaces, and features from different domain paths could be complementary. Therefore, the fused features also bring the good performance in the supervised case, which will be verified in the experiment. 

Besides, our method is designed for the ``A+B$\rightarrow$C'' task, where ``A'', ``B'' and ``C'' denote different domains, and there are no available labels in ``A'' and ``B''. Thus, we can directly convert the original task to the ``A+B$\rightarrow$A, B'' task, which is similar to the UDA task, yet removing the labels of the source domain during training. We call this task as UDAw/oSL (\ie, UDA without source labels). Particularly, if using our method to solve this UDAw/oSL task, we can obtain the competitive results when compared to the strong UDA-ReID baselines, as validated in the experiment.

\section{Experiments}\label{s-experiment}
In this part, we firstly introduce the experimental datasets and settings in Section~\ref{sec:EXP-DS}. Then, we compare the proposed method with the basic networks in Section~\ref{sec:EXP-CUA}, respectively. To validate the effectiveness of various components in the proposed framework, we conduct ablation studies in Section~\ref{sec:EXP-SS}. Besides, we further analyze the property of the proposed method in Section~\ref{sec:EXP-FA}. In Section~\ref{sec:EXP-ESDC}, we also validate the effectiveness of our method in the supervised DG-ReID. Lastly, we convert UDG-ReID to the UDAw/oSL-ReID task in Section~\ref{sec:EXP-UDA}, which shows the non-necessity of source labels in the typical UDA-ReID task.

\subsection{Dataset and Setting}~\label{sec:EXP-DS}
\textbf{Datasets.}  We evaluate our method on three large-scale image datasets: Market1501 (\textbf{Ma})~\cite{DBLP:conf/iccv/ZhengSTWWT15}, DukeMTMC-reID (\textbf{D})~\cite{DBLP:conf/iccv/ZhengZY17}, MSMT17 (\textbf{Ms})~\cite{DBLP:conf/cvpr/WeiZ0018}, and CUHK03-NP (\textbf{C})~\cite{DBLP:conf/cvpr/ZhongZCL17,DBLP:conf/cvpr/LiZXW14}. The partition of training and testing sets on each dataset is the same as~\cite{DBLP:conf/cvpr/ZhaoZYLLLS21}.  
 \textbf{Market1501 (Ma)} contains 1,501 persons with 32,668 images from six cameras. Among them, $12,936$ images of $751$ identities are used as training set. For evaluation, there are $3,368$ and $19,732$ images in the query set and the gallery set, respectively. \textbf{DukeMTMC-reID (D)} has $1,404$ persons from eight cameras, with $16,522$ training images, $2,228$ queries, and $17,661$ gallery images.
 \textbf{MSMT17 (Ms)} is collected from a 15-camera network deployed on campus. The training set contains $32,621$ images of $1,041$ identities. For evaluation, $11,659$ and $82,161$ images are adopted as query and gallery images, respectively. 
 \textbf{CUHK03-NP (C)} has an average of 4.8 images in each camera. The dataset provides both manually labeled bounding boxes and DPM-detected bounding boxes. The dataset provides both manually labeled bounding boxes and DPM-detected bounding boxes. On this dataset, there are $7,365$ training images, and $1,400$ images and $5,332$ images in query set and gallery set are used in the testing stage.
For all datasets, we employ CMC accuracy and mAP for ReID evaluation~\cite{DBLP:conf/iccv/ZhengSTWWT15,DBLP:journals/pami/LiZG20}.

\textbf{Implementation Details.} 
In this experiment, we use the pre-trained ResNet-50 on ImageNet~\cite{DBLP:conf/cvpr/DengDSLL009} to initialize the network parameters. In a batch, the number of IDs and the number of images per person are set as $16$ and $4$ to produce triplets, respectively.
 The learning rate is $3.5\times 10^{-4}$. The proposed model is trained with the Adam optimizer in a total of $50$ epochs. The size of the input image is $256 \times 128$. For data augmentation, we merely perform random cropping and
random flipping. Similar to~\cite{DBLP:conf/cvpr/JinLZ0Z20}, we do not utilize random erasing because it will degenerate the cross-domain ReID performance. In all experiments, the ``\textbf{Baseline}''  denotes using the raw ResNet-50~\cite{DBLP:conf/cvpr/HeZRS16} in Fig.~\ref{fig07}.
 Particularly, all experiments on all datasets utilize the same setting.

\begin{table}[t]
  \centering
  \caption{Experimental results of different methods on Market1501, DukeMTMC-reID and MSMT17. ``Supervised'' denotes giving all source labels for our method (DSAF). The bold is the best result.}
    \begin{tabular}{c|cccc}
    \toprule
    \multirow{2}[2]{*}{Method} & \multicolumn{4}{c}{Market+Duke$\rightarrow$MSMT} \\
\cmidrule{2-5}          & mAP   & Rank-1 & Rank-5 & Rank-10 \\
    \midrule
    Baseline & 3.7   & 11.6  & 19.5  & 23.5 \\
    IBN-Net~\cite{DBLP:conf/eccv/PanLST18} & 5.7   & 17.4  & 26.7  & 31.8 \\
    DSAF  & \textbf{10.3} & \textbf{28.9} & \textbf{39.9} & \textbf{45.5} \\
     \midrule
    Supervised & 11.7  & 31.1  & 43.3  & 48.9 \\
    \midrule
          & \multicolumn{4}{c}{Market+MSMT$\rightarrow$Duke} \\
    \midrule
    Baseline & 15.3  & 26.6  & 41.9  & 48.3 \\
    IBN-Net~\cite{DBLP:conf/eccv/PanLST18} & 18.2  & 34.1  & 49.4  & 55.9 \\
    DSAF  & \textbf{39.9} & \textbf{61.3} & \textbf{73.4} & \textbf{77.5} \\
    \midrule
    Supervised & 44.3  & 63.7 & 76.1  & 80.4 \\
    \midrule
          & \multicolumn{4}{c}{Duke+MSMT$\rightarrow$Market} \\
    \midrule
    Baseline & 21.3  & 47.0    & 64.1  & 70.9 \\
    IBN-Net~\cite{DBLP:conf/eccv/PanLST18} & 25.5  & 54.1  & 69.9  & 76.5 \\
    DSAF  & \textbf{37.6} & \textbf{68.2} & \textbf{81.4} & \textbf{86.4} \\
    \midrule
    Supervised & 40.9  & 70.9  & 83.9  & 88.1 \\
    \bottomrule
    \end{tabular}%
  \label{tab01}%
\end{table}%

\begin{figure}
\centering
\subfigure[Ms+Ma$\rightarrow$D]{
\includegraphics[width=4.5cm]{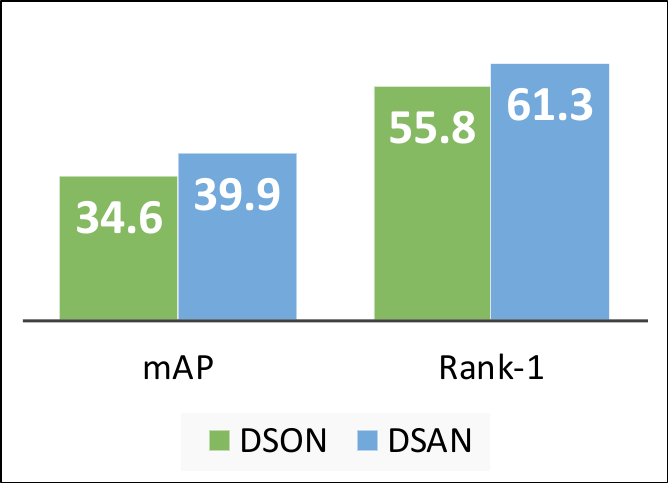}
} 
\subfigure[Ms+D$\rightarrow$Ma]{
\includegraphics[width=4.5cm]{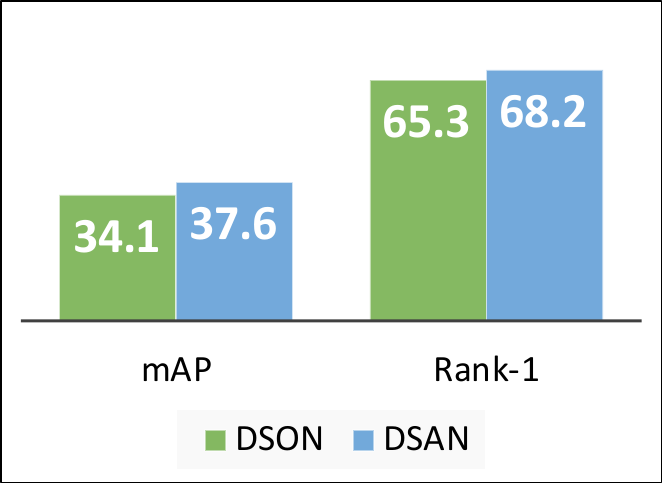}
}
\caption{Comparison between DSAN and DSON~\cite{DBLP:conf/eccv/SeoSKKHH20}.}
\label{fig08}
\end{figure}


\subsection{Comparison with Basic Networks}~\label{sec:EXP-CUA}
In this part, we conduct the experiment to compare our method with some basic networks, \eg, ResNet-50~\cite{DBLP:conf/cvpr/HeZRS16} and IBN-Net~\cite{DBLP:conf/eccv/PanLST18}, as reported in Tab.~\ref{tab01}. 
As seen, IBN-Net outperforms the baseline (\ie, ResNet-50) model, which mainly thanks to the IN structure in the network, thus it can reach better generalization ability in the unseen target domain. Besides, our method can achieve better performance than other methods. For example, compared to IBN-Net, in the ``Duke+MSMT$\rightarrow$Market'' task, our method improves mAP and Rank-1 by +12.1\% (37.6 vs. 25.5) and +14.1\% (68.2 vs. 54.1). This main reason is that, in the UDG-ReID task we need to simultaneously solve the domain interference during pseudo-labeling and the generalization issue in unseen target domains. However, IBN-Net merely considers IN to deal with the latter issue. 

Moreover, we report the experimental results of our method when giving the labels for all training samples. As seen in Tab.~\ref{tab01}, our method in the UDG-ReID task slightly drops when compared to the supervised DG setting. Particularly, our method in the supervised DG setting also can achieve competitive performance when compared to some SOTA methods, as shown in Sec.~\ref{sec:EXP-ESDC}.

In addition, we also compare the proposed DSAN with DSON~\cite{DBLP:conf/eccv/SeoSKKHH20}, which utilizes the domain-specific normalization technique for the supervised DG classification task, and it can also be employed to implement our DSAF. DSON conducts the fusion of the batch and instance statistics and then performs the normalization, which could be not robust for the unseen domains due to the inaccurate pseudo-labels of source domains in our unsupervised DG-ReID. As seen in Fig.~\ref{fig08}, our DSAN can outperform the DSON.

\subsection{Ablation Study}~\label{sec:EXP-SS}
In this part, we conduct the ablation study to validate the effectiveness of each component of our method, as reported in Tab.~\ref{tab02}. Firstly, based on the baseline model (\ie, the original ResNet-50), we employ the domain-specific normalization (DSBN) to alleviate the inter-domain interference, which can produce the more reliable pseudo-labels for unlabeled data. As seen in Tab.~\ref{tab02}, using DSBN indeed brings the improvement of performance in the target domain. Secondly, when we add the proposed DSAN into the above model, the mAP and Rank can be further boosted, which mainly owes to the generalization ability of IN in the proposed DSAN. In the ``Market+MSMT$\rightarrow$Duke'' task, our DSAN can increase +9.3\% (39.9 vs. 30.6) and +11.5\% (61.3 and 49.8) on mAP and Rank-1, respectively.

\begin{table}[t]
  \centering
  \caption{Ablation study on Market1501, DukeMTMC-reID and MSMT17. Note the bold is the best result.}
    \begin{tabular}{c|cccc}
    \toprule
    \multirow{2}[2]{*}{Method} & \multicolumn{4}{c}{Market+Duke$\rightarrow$MSMT} \\
\cmidrule{2-5}          & mAP   & Rank-1 & Rank-5 & Rank-10 \\
    \midrule
    Baseline & 3.7   & 11.6  & 19.5  & 23.5 \\
    +DSBN  & 6.3   & 19.4  & 28.8  & 33.5 \\
    +DSAN  & \textbf{10.3} & \textbf{28.9} & \textbf{39.9} & \textbf{45.5} \\
    \midrule
          & \multicolumn{4}{c}{Market+MSMT$\rightarrow$Duke} \\
    \midrule
    Baseline & 15.3  & 26.6  & 41.9  & 48.3 \\
    +DSBN  & 30.6  & 49.8  & 64.6  & 69.3 \\
    +DSAN  & \textbf{39.9} & \textbf{61.3} & \textbf{73.4} & \textbf{77.5} \\
    \midrule
          & \multicolumn{4}{c}{Duke+MSMT$\rightarrow$Market} \\
    \midrule
    Baseline & 21.3  & 47.0  & 64.1  & 70.9 \\
    +DSBN  & 29.9  & 60.7  & 75.8  & 81.1 \\
    +DSAN  & \textbf{37.6} & \textbf{68.2} & \textbf{81.4} & \textbf{86.4} \\
    \bottomrule
    \end{tabular}%
  \label{tab02}%
\end{table}%

\begin{figure}
\centering
\includegraphics[width=9cm]{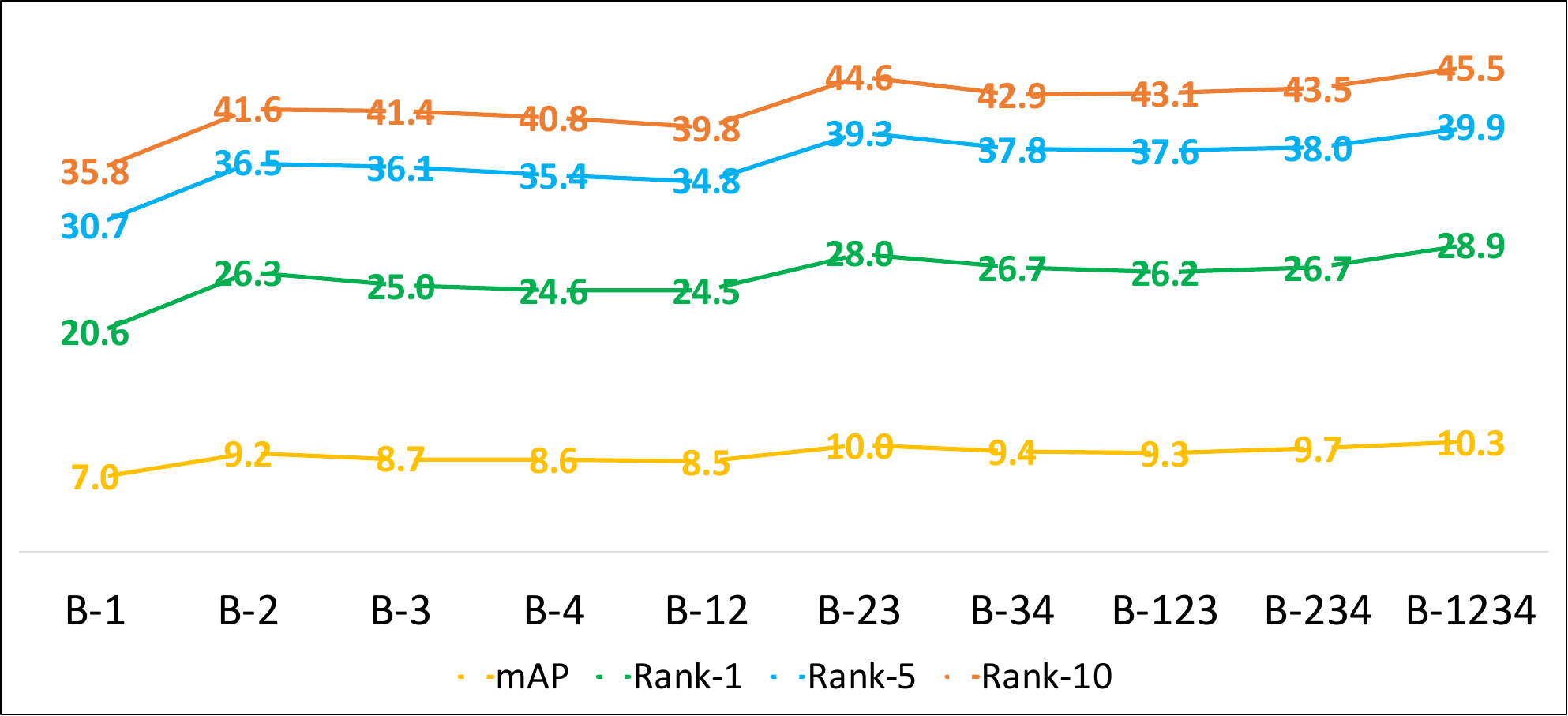}
 \caption{Results of different positions of DSAN in the backbone in the ``Market+Duke$\rightarrow$MSMT'' task. ``B-123'' denotes that we use the DSAN in Block 1, 2 and 3 of ResNet-50.}
\label{fig01}
\end{figure}

Besides, we validate the necessity of the affine transformation of IN in DSAN, \ie, $\gamma^{in},\beta^{in}$ in Eq.~(\ref{eq01}), as reported in Tab.~\ref{tab03}. As seen, if we remove the affine operation, the performance will have a large degradation, because using affine could make the feature maps of IN suitable for another half the feature maps of BN. Moreover, we verify the effectiveness of sharing affine of IN for all source domains. As seen in Tab.~\ref{tab03}, sharing affine operation can obtain better experimental results in the unseen target domains, and it is used in our all experiments. For example, in the ``Market+MSMT$\rightarrow$Duke'' task, using the independent affine operation of IN for each domain in DSAN will decrease -1.7\% (38.2 vs. 39.9) and -1.5\% (59.8 vs. 61.3) on mAP and Rank-1 when compared with the model using shared affine operation for all domains. This main reason is that, if we leverage the independent affine operation for each source domain, it could cause that the learnable affine sightly overfits to the corresponding domain. 

\begin{table}[t]
  \centering
  \caption{Experimental results with or without affine of IN in DSAN. ``Share'' denotes that if sharing the affine for each source domains. `` $\checkmark$ - $\checkmark$ '' is that DSAN includes the affine of IN and shares the unique affine for each domains in the training stage.}
    \begin{tabular}{c|cccc}
    \toprule
    \multirow{2}[2]{*}{Affine-Share} & \multicolumn{4}{c}{Duke+MSMT$\rightarrow$Market} \\
\cmidrule{2-5}          & mAP   & Rank-1 & Rank-5 & Rank-10 \\
    \midrule
      \xmark - \xmark    & 28.6  & 57.6  & 73.9  & 79.4 \\
        $\checkmark$ - \xmark   & 36.4  & 67.3  & 80.7  & 85.5 \\
       $\checkmark$ - $\checkmark$    & \textbf{37.6} & \textbf{68.2} & \textbf{81.4} & \textbf{86.4} \\
    \midrule
         & \multicolumn{4}{c}{Market+MSMT$\rightarrow$Duke} \\
    \midrule
          \xmark - \xmark  & 25.8  & 43.4  & 57.9  & 63.8 \\
         $\checkmark$ - \xmark  & 38.2  & 59.8  & 71.9  & 77.0 \\
         $\checkmark$ - $\checkmark$  & \textbf{39.9} & \textbf{61.3} & \textbf{73.4} & \textbf{77.5} \\
    \midrule
          & \multicolumn{4}{c}{Market+Duke$\rightarrow$MSMT} \\
    \midrule
         \xmark - \xmark & 6.0   & 18.1  & 27.4  & 32.4 \\
         $\checkmark$ - \xmark & 9.7   & 27.8  & 39.3  & 44.9 \\
        $\checkmark$ - $\checkmark$   & \textbf{10.3} & \textbf{28.9} & \textbf{39.9} & \textbf{45.5} \\
    \bottomrule
    \end{tabular}%
  \label{tab03}%
\end{table}%


\subsection{Further Analysis}~\label{sec:EXP-FA}
\textbf{Evaluation on the policy of splitting the channels into two parts.} We conduct the experiment of different policies on splitting the channels into two parts, as listed in Tab.~\ref{tab_r1}. We split the channel into 25\% (50\%, or 75\%) IN and 75\% (50\%, or 25\%) BN. As seen, when we split the channel into 50\% IN and 50\% BN, the performance is best, which reveals that the domain-specific information based on BN and sample-specific information based on IN is equally important in our task. In our all experiments, we split the channel into 50\% IN and 50\% BN.
\begin{table}[t]
  \centering
  \caption{Experimental results of different policies on splitting the channels into two parts. In this table, we conduct an experiment on Market, Duke and MSMT. We use two datasets to train the model and test the remaining dataset. For example, ``Market'' denotes that the model is trained on MSMT and Duke and tested on Market. }
    \begin{tabular}{cc|cc|cc|cc}
    \toprule
    \multirow{2}[4]{*}{IN} & \multirow{2}[4]{*}{BN} & \multicolumn{2}{c|}{Market} & \multicolumn{2}{c|}{Duke} & \multicolumn{2}{c}{MSMT} \\
\cmidrule{3-8}          &       & mAP   & Rank-1 & mAP   & Rank-1 & mAP   & Rank-1 \\
    \midrule
    0.25  & 0.75  & 36.3  & 67.1  & 36.6  & 56.3  & 9.3   & 26.2 \\
    0.75  & 0.25  & 36.3  & 65.6  & 38.5  & 58.3  & 10.3  & 28.2 \\
    0.50  & 0.50  & 37.6  & 68.2  & 39.9  & 61.3  & 10.3  & 28.9 \\
    \bottomrule
    \end{tabular}%
  \label{tab_r1}%
\end{table}%

\textbf{Evaluation on the position of DSAN.} As known, ResNet-50 mainly consists of four blocks~\cite{DBLP:conf/cvpr/HeZRS16}. Here, we evaluate the impact of the different positions of DSAN in the backbone, as shown in Fig.~\ref{fig01}. Compared to the baseline method in the ``Market+Duke$\rightarrow$MSMT'' task as shown in Tab.~\ref{tab02} (\ie, mAP=3.7\% and Rank-1=11.6\%), the proposed DSAN used in any position can reach the better performance. Besides, we observe that, if using DSAN in a Block, the model with DSAN in Block 2 has better results than the model with DSAN in other Blocks. Moreover, when we use the proposed DSAN in all Blocks of the backbone, the best results can be yielded. In particular, using DSAN does not bring more parameters than the original ResNet-50. In our experiment, we use the same setting in all tasks.

\textbf{Evaluation on more source domains.}  
We evaluate our method in more source domains, as reported in Tab.~\ref{tab04}. As seen, the model trained via three source domains can obtain better performance than the model trained with two source domains. For example, the mAP and Rank-1 of the model in the Ma+D+Ms$\rightarrow$C task outperform the model in the ``Ma+Ms$\rightarrow$C'' task by +2.1\% (17.0 vs. 14.9) and +2.0\% (16.9 vs. 14.9). This shows that our method can also effectively alleviate the interference of the domain gap across different domains when there are more available unlabeled source domains in the training stage.

\begin{figure}
\centering
\subfigure[Performance]{
\includegraphics[width=5cm]{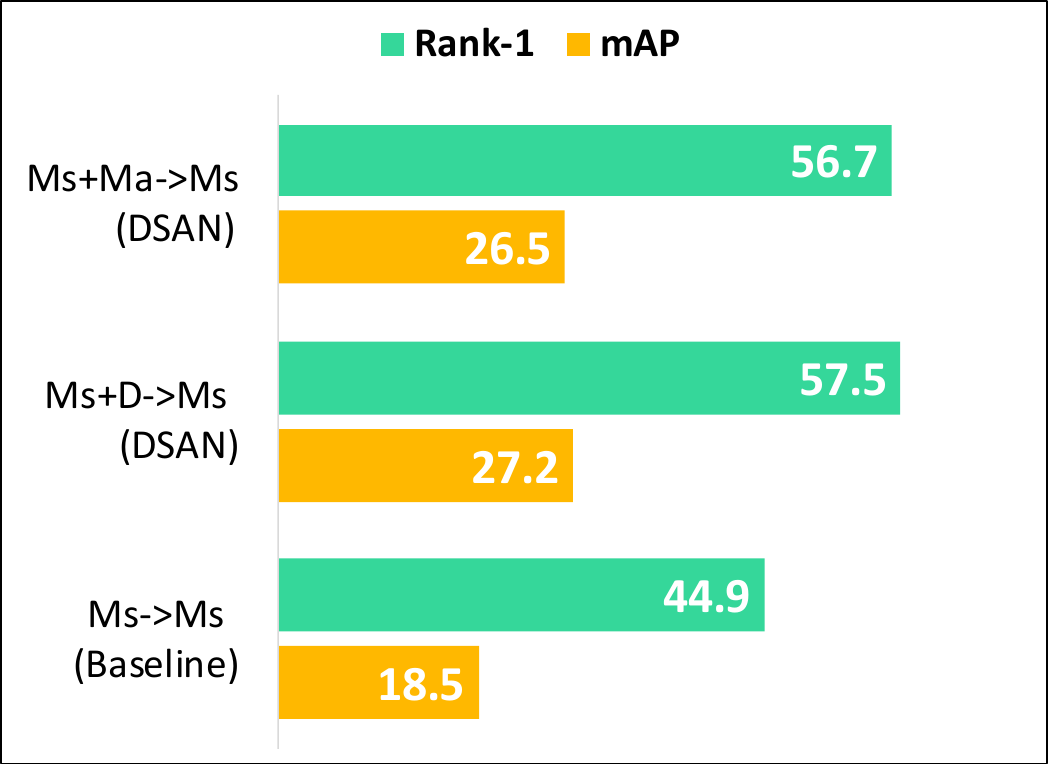}
} 
\subfigure[Clustering evaluation]{
\includegraphics[width=5cm]{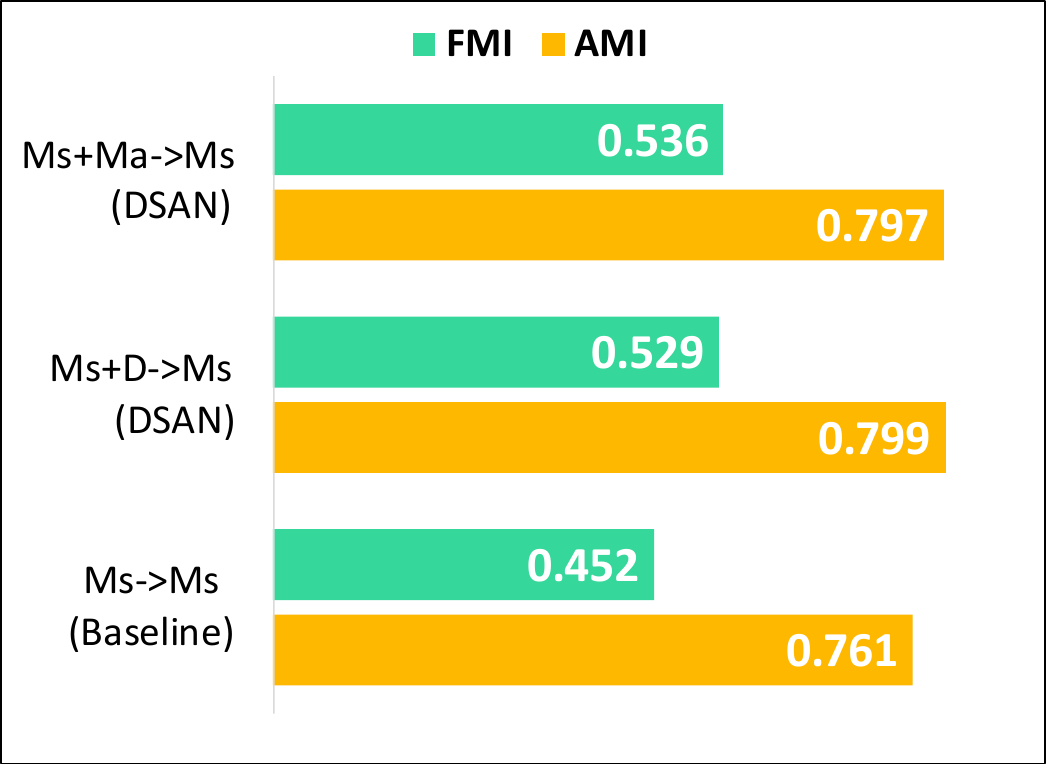}
}
  \caption{Performance and clustering results on MSMT17. As seen in this figure, our method can effectively mitigate the existing issue as shown in Fig.~\ref{fig02}.}
\label{fig03}
\end{figure}


\begin{table}[t]
  \centering
  \caption{Experimental results of the model trained using different numbers of source domains and tested on CUHK03-NP.}
    \begin{tabular}{c|cccc}
    \toprule
    Setting  & mAP   & Rank-1 & Rank-5 & Rank-10 \\
    \midrule
    Ma+D$\rightarrow$C & 12.7  & 13.1  & 23.8  & 29.9  \\
    D+Ms$\rightarrow$C & 12.1  & 12.1  & 21.6  & 27.7  \\
    Ma+Ms$\rightarrow$C & 14.9  & 14.9  & 27.1  & 34.8  \\
    Ma+D+Ms$\rightarrow$C & \textbf{17.0} & \textbf{16.9} & \textbf{29.3} & \textbf{37.3} \\
    \bottomrule
    \end{tabular}%
  \label{tab04}%
\end{table}

\textbf{Evaluation on the pseudo-label quality.}  
As aforementioned, if we use the conventional baseline method to conduct the proposed UDG-ReID, it will produce poor pseudo-labels for the unlabeled training domains due to the domain-shift between different source domains. In this part, we validate the efficacy of our method from two different perspectives, including performance and clustering accuracy in the source domains, as shown in Tab.~\ref{tab06} and Fig.~\ref{fig03}. Firstly, as seen in Tab.~\ref{tab06}, our method can produce better results than the baseline method, and the clustering accuracy is higher than the baseline. In the ``Duke+MSMT$\rightarrow$Market'' task, the mAP and Rank-1 can improve +17\% (27.2 vs. 10.2) and +31.0\% (57.5 vs. 26.5), and AMI and FMI can increase +0.142 (0.799 vs. 0.657) and +0.229 (0.529 vs. 0.300). This confirms that our method can alleviate the inter-domain interference during generating pseudo-labels so as to yield more reliable pseudo-labels to train the model.

Moreover, we also compare our method with the model trained in a single domain. If we utilize multiple domains to train the baseline model, it will result in poor performance when compared to employing a single domain to train the model, as shown in Fig.~\ref{fig02}. Differently, as seen in  Fig.~\ref{fig03}, our method can alleviate the issue, and can fully exploit each domain to boost the discrimination of the model.
\begin{table}[htbp]
  \centering
  \caption{Experimental results and clustering accuracy of source domains in different tasks. AMI (Adjusted Mutual Information) and FMI (Fowlkes-Mallows Index) are two clustering evaluation protocols, and the larger value is better.}
    \resizebox{\linewidth}{!}{
    \begin{tabular}{c|cccc|cc|cccc|cc}
    \toprule
    \multirow{2}[6]{*}{Method} & \multicolumn{12}{c}{Train: Market + Duke} \\
\cmidrule{2-13}          & \multicolumn{6}{c|}{Test: Market}              & \multicolumn{6}{c}{Test: Duke} \\
\cmidrule{2-13}          & mAP   & rank-1 & rank-5 & rank-10 & AMI   & FMI   & mAP   & Rank-1 & Rank-5 & Rank-10 & AMI   & FMI \\
    \midrule
    Baseline & 48.1  & 72.1  & 86.6  & 90.4  & 0.781 & 0.449 & 41.0  & 57.7  & 71.8  & 76.7  & 0.796 & 0.508 \\
    DSAF  & \textbf{69.9} & \textbf{88.7} & \textbf{95.0} & \textbf{96.5} & \textbf{0.874} & \textbf{0.646} & \textbf{55.7} & \textbf{73.8} & \textbf{83.9} & \textbf{87.5} & \textbf{0.845} & \textbf{0.585} \\
    \midrule
    \multicolumn{1}{r}{} & \multicolumn{12}{c}{Train: Market + MSMT} \\
    \midrule
          & \multicolumn{6}{c|}{Test: Market}             & \multicolumn{6}{c}{Test: MSMT} \\
    \midrule
    Baseline & 44.4  & 69.8  & 84.5  & 89.4  & 0.773 & 0.464 & 8.4   & 22.5  & 32.9  & 37.7  & 0.627 & 0.242 \\
    DSAF  & \textbf{72.8} & \textbf{89.5} & \textbf{95.8} & \textbf{97.4} & \textbf{0.886} & \textbf{0.689} & \textbf{26.5} & \textbf{56.7} & \textbf{68.7} & \textbf{73.6} & \textbf{0.797} & \textbf{0.536} \\
    \midrule
    \multicolumn{1}{r}{} & \multicolumn{12}{c}{Train: Duke + MSMT} \\
    \midrule
          & \multicolumn{6}{c|}{Test: Duke}               & \multicolumn{6}{c}{Test: MSMT} \\
          \midrule
    Baseline & 39.2  & 56.9  & 72.1  & 77.1  & 0.795 & 0.516 & 10.2  & 26.5  & 38.0  & 43.5  & 0.657 & 0.300 \\
    DSAF  & \textbf{59.0} & \textbf{76.0} & \textbf{85.0} & \textbf{88.2} & \textbf{0.846} & \textbf{0.582} & \textbf{27.2} & \textbf{57.5} & \textbf{69.3} & \textbf{73.8} & \textbf{0.799} & \textbf{0.529} \\
    \bottomrule
    \end{tabular}%
    }
  \label{tab06}%
\end{table}%

\begin{figure}
\centering
\subfigure[Market1501]{
\includegraphics[width=5cm]{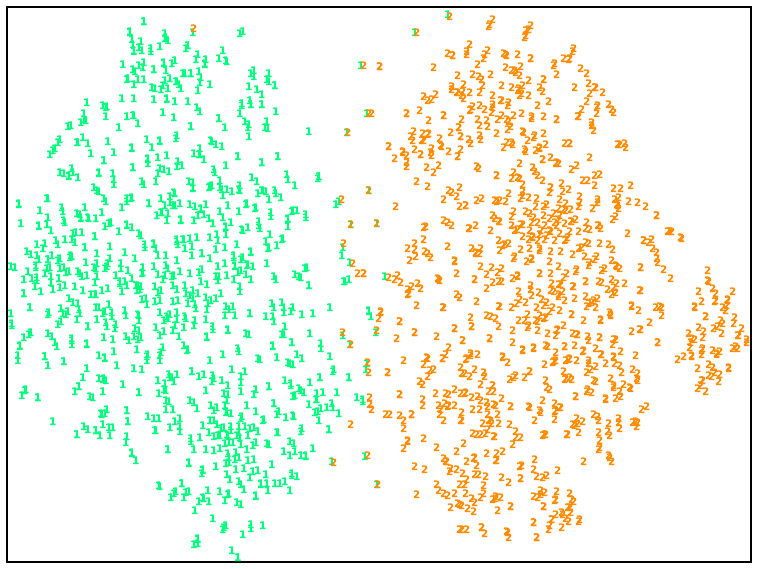}
} 
\subfigure[DukeMTMC-reID]{
\includegraphics[width=5cm]{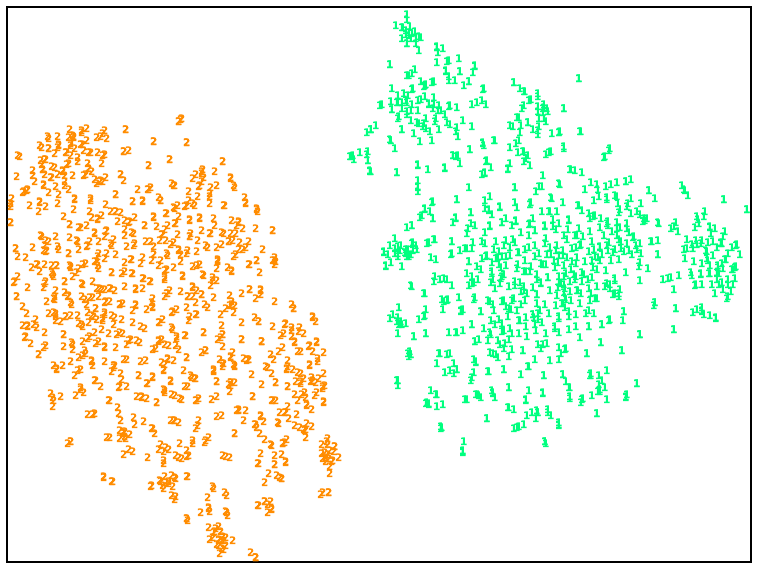}
}
\caption{Visualization of features from different paths of our model via t-SNE~\cite{van2008visualizing} on Market1501 and DukeMTMC-reID, which corresponds to the  ``Duke+MSMT$\rightarrow$Market'' and ``Market+MSMT$\rightarrow$Duke'' tasks. Note that, in each figure, different colors denote that the same samples from the same dataset pass different paths.}
\label{fig06}
\end{figure}

\textbf{Effectiveness of feature fusion in the testing stage.} 
Since our DSAN uses the independent BN for each domain, we fuse the feature from each branch to form the final feature for testing. As seen in Tab.~\ref{tab07}, the fused features can yield better performance than the features from the single path. For example, in the ``Duke+MSMT$\rightarrow$Market'' task, using the fused feature can increase +3.1\% (37.6 vs. 34.5) and +2.6\% (37.6 vs. 35.0) when compared to the features from ``path-duke'' and ``path-msmt''. This main reason is that the features in each branch could be complementary due to the domain discrepancy, thus the fused features can bring further improvement of the performance. Here we also give the data distribution from each path, as shown in Fig.~\ref{fig06}. As seen, the features extracted by different paths are scattered. Thus, features of different domain-paths might be complementary, and fused features are more discriminative.

\begin{table}[htbp]
  \centering
  \caption{Comparison of the domain-path feature and the fused feature. ``Path-duke'' is the feature from Duke-path.}
    \begin{tabular}{c|cccc}
    \toprule
    \multirow{2}[2]{*}{Method} & \multicolumn{4}{c}{Market+Duke$\rightarrow$MSMT} \\
\cmidrule{2-5}          & mAP   & Rank-1 & Rank-5 & Rank-10 \\
    \midrule
    Path-market & 8.8   & 25.0  & 36.4  & 41.8  \\
    Path-duke & 9.4   & 27.3  & 38.5  & 43.9  \\
    Fusion & \textbf{10.3} & \textbf{28.9} & \textbf{39.9} & \textbf{45.5} \\
    \midrule
          & \multicolumn{4}{c}{Market+MSMT$\rightarrow$Duke} \\
    \midrule
    Path-market & 32.5  & 52.9  & 67.6  & 72.5  \\
    Path-msmt & 38.3  & 60.5  & 72.9  & 76.8  \\
    Fusion & \textbf{39.9} & \textbf{61.3} & \textbf{73.4} & \textbf{77.5} \\
    \midrule
          & \multicolumn{4}{c}{Duke+MSMT$\rightarrow$Market} \\
    \midrule
    Path-duke & 35.0  & 65.4  & 79.4  & 85.2  \\
    Path-msmt & 34.5  & 65.1  & 79.2  & 84.5  \\
    Fusion & \textbf{37.6} & \textbf{68.2} & \textbf{81.4} & \textbf{86.4} \\
    \bottomrule
    \end{tabular}%
  \label{tab07}%
\end{table}%


\subsection{Extension to Supervised DG-ReID Case}
\label{sec:EXP-ESDC}
In this experiment, we also use our method under the supervised DG person ReID, and compare it with the recent DG-ReID methods, such as QAConv50 and  M$^3$L,  as reported in Tab.~\ref{tab08}.  QAConv50~\cite{DBLP:conf/eccv/LiaoS20}  treats image matching as finding local correspondences in feature maps, and constructs query-adaptive convolution kernels on the fly to
achieve local matching. 
 M$^3$L~\cite{DBLP:conf/cvpr/ZhaoZYLLLS21} introduces a meta-learning strategy to simulate the train-test process of domain generalization for learning more generalizable models, the memory-based identification loss to overcome the unstable meta-optimization caused by the parametric classifier, and a meta batch normalization layer to diversify meta-test features.
As seen in Tab.~\ref{tab08}, our method outperforms these methods, despite our method being very simple. Particularly, on the large-scale dataset (MSMT17), the Rank-1 of our method improves the results of M$^3$L by +5.3\%  (42.4 vs. 37.1), which owes to the generalization capability of the fused features, as reported in Tab.~\ref{tab08}.

\begin{table}[htbp]
  \centering
  \caption{Experimental results under the supervised DG setting. When we evaluates models on Market1501 (or MSMT17), ``DSAF-P1'', ``DSAF-P2'' and ``DSAF-P3'' indicate using the features from the Path-Duke, Path-MSMT17 and Path-CUHK03 (or Path-Market, Path-Duke and Path-CUHK03), respectively.}
    \begin{tabular}{c|cc|cc}
    \toprule
    \multirow{2}[2]{*}{Method} & \multicolumn{2}{c|}{D+Ms+C$\rightarrow$Ma} & \multicolumn{2}{c}{Ma+D+C$\rightarrow$Ms} \\
\cmidrule{2-5}          & mAP   & Rank-1 & mAP   & Rank-1 \\
    \midrule
    QAConv50~\cite{DBLP:conf/eccv/LiaoS20}  & 39.5  & 68.6  & 10.0  & 29.0  \\
      M$^3$L\small{(ResNet50)}~\cite{DBLP:conf/cvpr/ZhaoZYLLLS21}    & 51.1  & 76.5  & 13.1  & 32.0  \\
     M$^3$L\small{(IBN-Net50)}~\cite{DBLP:conf/cvpr/ZhaoZYLLLS21}    & 52.5  & 78.3  & 15.4  & 37.1  \\
       \midrule
       DSAF-P1 & 44.4  & 73.6  & 10.8  & 29.7  \\
       DSAF-P2  & 45.3  & 73.5  & 14.9  & 39.3  \\
       DSAF-P3  &  44.2 & 72.7  & 9.8  & 29.7  \\
    DSAF (ours) & \textbf{53.2} & \textbf{79.7} & \textbf{16.8} & \textbf{42.4} \\
    \bottomrule
    \end{tabular}%
  \label{tab08}%
\end{table}%
\subsection{UDA-ReID without Source Labels}~\label{sec:EXP-UDA}
As mentioned in Sec.~\ref{sec:DNF}, the proposed UDG-ReID task can be converted to the UDAw/oSL-ReID task, which equals to the case that UDA removes labels from the source domain. The UDAw/oSL-ReID task does not need to annotate the originally collected data, thus it is more valuable than the UDA task in the real-world application. We conduct the experiment to compare our method in the UDAw/oSL-ReID task with strong UDA-ReID baselines, such as SBase-1~\cite{DBLP:conf/aaai/ZhengLZZZ21}  and SBase-2~\cite{Dai_2021_ICCV}, as reported in Tab.~\ref{tab09}. Particularly, in this experiment, we add the random erasing data augmentation~\cite{DBLP:conf/aaai/Zhong0KL020} to enrich the diversity of samples due to no domain gap between the training set and the testing set, which is also used in SBase-1~\cite{DBLP:conf/aaai/ZhengLZZZ21}  and SBase-2~\cite{DBLP:conf/cvpr/WangZHS20}. Besides, SBase-1 utilizes contrastive loss across memory bank~\cite{DBLP:conf/cvpr/WuXYL18} and mean teacher method~\cite{DBLP:conf/nips/TarvainenV17}, and SBase-2~\cite{Dai_2021_ICCV} uses XBM  to mine more hard negatives for the triplet loss, which is a variant of the memory bank. Differently, our method does not leverage these above operations.  As seen in Tab.~\ref{tab09}, although there is no available label information in the source domain, the UDAw/oSL-ReID  task with our method can obtain competitive results when compared to strong baseline UDA methods. Furthermore, we also conduct the UDA case with our method, and the results are also similar to the UDAw/oSL-ReID case. This means that removing the labels from the source domain does not bring the largely negative impact in the UDA task. Thus, the UDAw/oSL-ReID might be more valuable than the UDA-ReID in future ReID community.
\begin{table}[htbp]
  \centering
  \caption{Experimental results of different methods in the UDA-ReID task and the UDAw/oSL-ReID task, respectively.}
    \begin{tabular}{c|cc|cc}
    \toprule
    \multirow{2}[1]{*}{Task} & \multicolumn{2}{c|}{Ma+D$\rightarrow$Ma} & \multicolumn{2}{c}{Ma+D$\rightarrow$D} \\
\cmidrule{2-5}          & mAP   & Rank-1 & mAP   & Rank-1 \\
    \midrule
    UDA\small{(SBase-1)}~\cite{DBLP:conf/aaai/ZhengLZZZ21} & 75.4 & 89.8 &64.8 & 79.7 \\
    UDA\small{(SBase-2)}~\cite{Dai_2021_ICCV} & 79.1  & 91.2  & 65.8  & 80.1 \\
     UDA\small{(DSAF)} & 78.6  & 91.7  & 65.2  & 80.1 \\
    \midrule
    UDAw/oSL\small{(DSAF)}   & 78.5  & 91.0  & 64.2  & 78.6 \\
    \bottomrule
    \end{tabular}%
  \label{tab09}%
\end{table}%

\section{Conclusion}\label{s-conclusion}
 This paper proposes an unsupervised DG-ReID task, where there is no available label information in all source domains. It is more challenging than the typical DG person ReID task. 
 Considering there exist two problems in the task, including the inter-domain interference during pseudo-labeling and the domain discrepancy between the source and target domains, we propose a unified method to simultaneously solve the two issues, called domain-specific adaptive framework (DSAF), which can reduce the impact of domain interference to produce the reliable pseudo-labels and enhance the robustness of the model in the unseen domain.
  We conduct extensive experiments on multiple benchmark datasets to confirm the efficacy of the proposed method. Moreover, we convert the UDG-ReID task to the UDAw/oSL-ReID task, which can obtain the competitive results using our method when compared to the strong UDA-ReID baselines.

\section{Acknowledgment}
This work was supported by NSFC Program (62206052, 62125602, 62076063, 62222604), CAAI-Huawei MindSpore Project (CAAIXSJLJJ-2021-042A), China Postdoctoral Science Foundation Project (2021M690609), Jiangsu Natural Science Foundation Project (BK20210224), and CCF-Lenovo Bule Ocean Research Fund.



\bibliographystyle{elsarticle-num}
\bibliography{egbib.bib}







\end{document}